\documentclass[journal,compsoc]{IEEEtran}
\usepackage[T1]{fontenc}

\usepackage{color}
\usepackage{array}
\usepackage{colortbl}

\usepackage{cite}

\ifCLASSINFOpdf
  \usepackage[pdftex]{graphicx}
  \graphicspath{{../pdf/}{../jpeg/}}
  \DeclareGraphicsExtensions{.pdf,.jpeg,.png}
\else
  \usepackage[dvips]{graphicx}
  \graphicspath{{../eps/}}
  \DeclareGraphicsExtensions{.eps}
\fi
\usepackage{amsmath}
\usepackage[cmintegrals]{newtxmath}

\usepackage{booktabs}
\usepackage{subfigure}
\usepackage{hyperref} 
\usepackage{cite}
\usepackage{mathrsfs}

\usepackage{array}
\usepackage{bbding}
\usepackage{ragged2e}

\begin{document}
%
\title{Facial Expression Recognition with Visual Transformers and Attentional Selective Fusion}
%
%
%

\author{Fuyan~Ma,
        Bin~Sun,~\IEEEmembership{Member,~IEEE,}
        and~Shutao~Li,~\IEEEmembership{Fellow,~IEEE}
\IEEEcompsocitemizethanks{
\IEEEcompsocthanksitem This work is supported by the National Key Research and Development Project (2018YFB1305200), the National Natural Science Fund of China (61801178) and the Key-Area Research and Development Plan of Guangdong Province (2018B010107001).
\IEEEcompsocthanksitem Fuyan Ma and Bin Sun are with College of Electrical and Information Engineering, and with the Key Laboratory of Visual Perception and Artificial Intelligence of Hunan Province, Hunan University, Changsha, 410082, China. (mafuyan@hnu.edu.cn; sunbin611@hnu.edu.cn)
\IEEEcompsocthanksitem Shutao Li is with College of Electrical and Information Engineering, with the State Key Laboratory of Advanced Design and Manufacturing for Vehicle Body and with the Key Laboratory of Visual Perception and Artificial Intelligence of Hunan Province, Hunan University, Changsha, 410082,China.(shutao\_li@hnu.edu.cn)
}
}

\IEEEtitleabstractindextext{
\justifying  
\begin{abstract}
Facial Expression Recognition (FER) in the wild is extremely challenging due to occlusions, variant head poses, face deformation and motion blur under unconstrained conditions.
Although substantial progresses have been made in automatic FER in the past few decades, previous studies were mainly designed for lab-controlled FER.
Real-world occlusions, variant head poses and other issues definitely increase the difficulty of FER on account of these information-deficient regions and complex backgrounds.
Different from previous pure CNNs based methods, we argue that it is feasible and practical to translate facial images into sequences of visual words and perform expression recognition from a global perspective.
Therefore, we propose the Visual Transformers with Feature Fusion (VTFF) to tackle FER in the wild by two main steps.
First, we propose the attentional selective fusion (ASF) for leveraging two kinds of feature maps generated by two-branch CNNs.
The ASF captures discriminative information by fusing multiple features with the global-local attention.
The fused feature maps are then flattened and projected into sequences of visual words.
Second, inspired by the success of Transformers in natural language processing, we propose to model relationships between these visual words with the global self-attention.
The proposed method is evaluated on three public in-the-wild facial expression datasets (RAF-DB, FERPlus and AffectNet).
Under the same settings, extensive experiments demonstrate that our method shows superior performance over other methods, setting new state of the art on RAF-DB with 88.14\%, FERPlus with 88.81\% and AffectNet with 61.85\%.
The cross-dataset evaluation on CK+ shows the promising generalization capability of the proposed method.
\end{abstract}

\begin{IEEEkeywords}
Facial expression recognition in the wild, global-local attention, Transformers, global self-attention.
\end{IEEEkeywords}}

\maketitle

\IEEEpeerreviewmaketitle

\section{Introduction\label{sec:introduction}}
%
%
%
%
\IEEEPARstart{U}{nderstanding} human emotional states is the fundamental task to develop emotional intelligence, which is an interdisciplinary field spanning from different research areas, such as psychology and computer science.
Facial expression is one of the most natural, powerful and universal signals for human beings to convey their emotional states and intentions\cite{darwin1998expression},\cite{tian2001recognizing}.
\begin{figure}[ht]
  \centering 
  \includegraphics[scale=0.42]{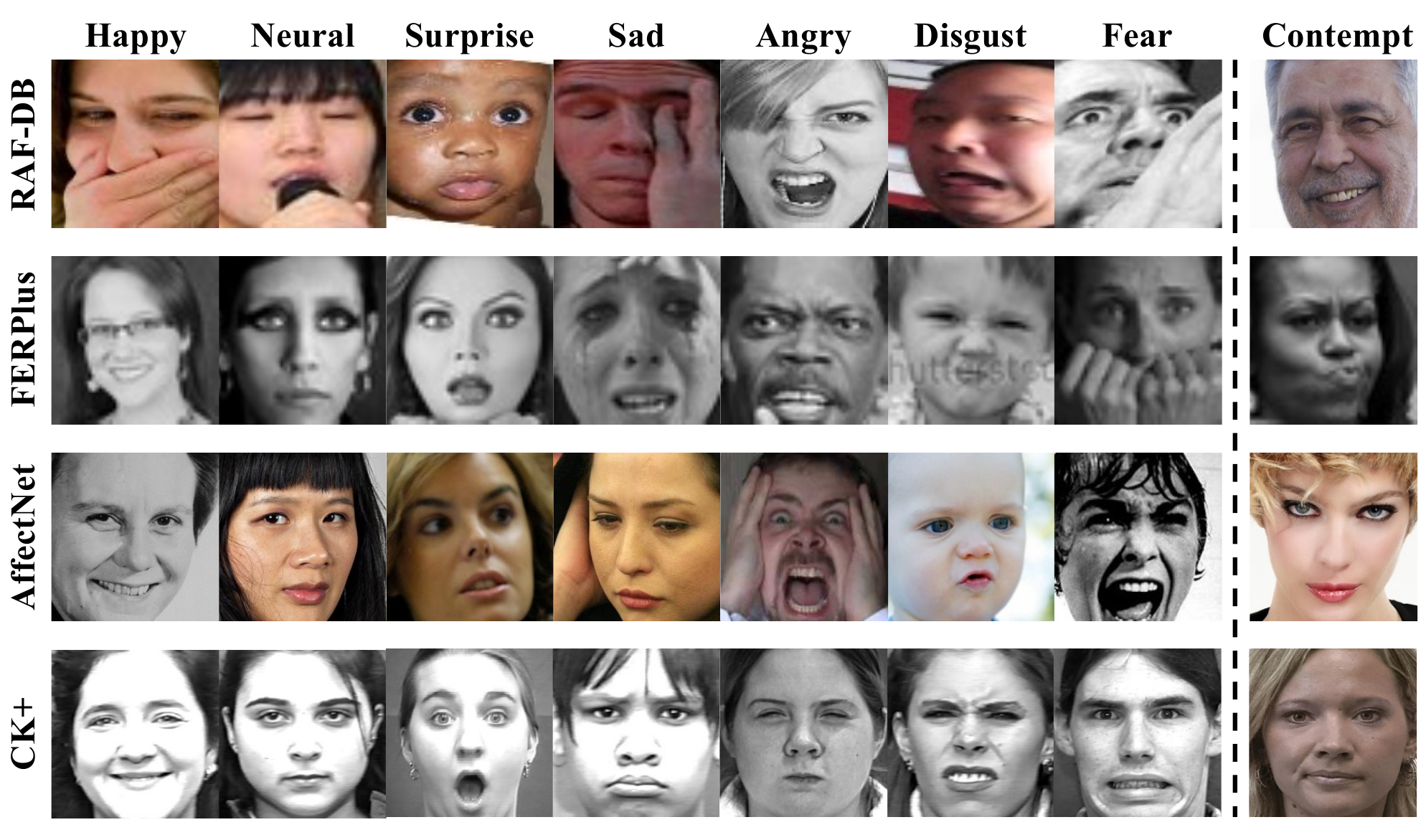} 
  \caption{The samples from RAF-DB, FERPlus, AffectNet and CK+.
  Variant head poses, occlusions and other unconstrained conditions can be seen in above images.
  Note that RAF-DB is annotated with seven basic expressions and FERPlus, AffectNet, CK+ are with eight expression labels, including the contempt category.
  }
  \label{samples}
\end{figure}
Facial expression recognition (FER) systems have various applications, including human-robot interaction (HRI), mental health assessment and driver fatigue monitoring.
Therefore, numerous research endeavours have been invested for promoting the development of FER.

As shown in Fig. \ref{samples}, the challenges of FER in the wild mainly come from occlusions, variant head poses, face deformation, motion blur, insufficient qualitative data, etc., which lead to significant changes of the facial appearance.
These unexpected issues definitely increase the difficulty of expression recognition.
With the advance of machine learning, especially deep learning, researchers have made great progress on FER in the past decades.
Before deep learning era, traditional FER methods have mainly used handcrafted features and shallow learning (e.g., Histograms of Oriented Gradients (HOGs)\cite{dalal2005histograms}, Local Binary Patterns (LBP)\cite{shan2005robust},\cite{feng2005facial}, non-negative matrix factorization (NMF)\cite{buciu2004application} and sparse representation\cite{lee2014intra}).
With the popularity of data-driven techniques, deep learning based methods have exceeded traditional methods by a large margin and achieved state-of-the-art FER performance (e.g.,\cite{li2017reliable},\cite{li2018occlusion},\cite{wang2020region},\cite{wang2020suppressing}).


Despite the significant success of learned representation for constrained FER, the performance of FER in the wild is still far from satisfactory. 
Majority of these proposed algorithms are implemented on lab-collected datasets, such as CK+\cite{lucey2010extended}, MMI\cite{valstar2010induced} and Oulu-CASIA \cite{zhao2011facial}.
These algorithms perform perfectly on these lab-collected FER datasets, because the controlled images are frontal with minimal illumination changes and limited occlusions.
However, the performance degrades dramatically on the real-world FER datasets, such as RAF-DB\cite{li2017reliable}, FERPlus\cite{barsoum2016training} and AffectNet\cite{mollahosseini2017affectnet}.
Compared with the real-world FER datasets, the number of images from the lab-collected FER datasets is relatively small.
Especially, convolutional neural networks (CNNs) \cite{lecun1989backpropagation} for the FER in the wild usually require sufficient training face images to ensure generalizability for real applications.
Most publicly available datasets for FER do not have a sufficient quantity of images for training.
Therefore, the performance of FER in the wild is limited by not only the unconstrained conditions but also the available data volume.

Different backgrounds, illuminations and head poses are fairly common under the unconstrained conditions, which are irrelevant to facial expressions\cite{li2020deep}.
Directly recognizing expression on such images is a big challenge.
To remove complex backgrounds and non-face areas, it is indispensable to detect faces before training the deep neural network to learn meaningful features.
Several face detectors like MTCNN\cite{zhang2016joint} and Dlib\cite{amos2016openface} are used to detect faces in complex scenarios.
After obtaining relatively accurate face regions, various methods have been proposed to improve the FER accuracy and enhance the generalization ability of expression recognition algorithms.
Most of the researchers design their methods for occlusion-aware \cite{bourel2001recognition,happy2014automatic} and multi-view FER, which are two main obstacles for FER in real-world scenarios.
Very recently, attention models have been successfully applied for FER to explore meaningful regions. 
Li et al. \cite{li2018patch} proposed a patch-gated CNN that integrates path-level attention for expression recognition with occlusion.
Besides, attention models have been successfully applied for FER to explore meaningful regions.
Similar to \cite{li2018patch}, several methods, such as \cite{wang2020region},\cite{fan2020facial},\cite{li2020attention}, also used attention-like mechanisms to focus on the most discriminative features to improve FER accuracy.
Non-frontal face images are overwhelmingly common in real-world scenarios.
Previous methods often treat expression recognition in non-frontal face images as the multi-view FER problem.
Zheng\cite{zheng2014multi} proposed to select the optimal sub-regions of a face that contribute most to the expression recognition based on a group sparse reduced-rank regression.
Liu et al. \cite{liu2018multi} proposed to tackle multi-view FER by three parts: the multi-channel feature extraction, the multi-scale feature fusion and the pose-aware recognition.
In addition, generative adversarial networks (GAN) have also been applied for multi-view FER.
GAN-based methods \cite{Zhang_2018_CVPR} \cite{lai2018emotion} can synthesize face images with large head pose variations to enlarge the training set for FER.
Especially, Zhang et al. \cite{zhang2020geometry} proposed an end to end model based on GAN for facial images synthesize with a set of facial landmarks and pose-invariant facial expression recognition by exploiting the geometry information.
Sun et al. \cite{sun2020unsupervised} proposed cyclic image generation for unsupervised cross-view facial expression recognition based on GAN.


Taking variant poses and occlusions in face images for example, we illustrate the motivation of our method.
Variant poses and the resize operation lead to face deformation compared with frontal faces, which can be seen as out-of-order subregions. 
Occlusions in face images result in information-deficient subregions.
By analogy with natural language, FER in the wild can be tackled in a different way.
There is an interesting phenomenon that we can understand the sentences, i.e., “\emph{I cdn’uolt blveiee taht I cluod aulaclty undresatnd what I was rdanieg: the phaonmneel pweor of the hmuan mnid.}”
One of the most possible reasons is that the cognitive mode of us human beings usually impels ourselves to look and think about this sentence globally.
Also like the image description task in Natural Language Processing (NLP), we may use several visual words to describe the emotional state for an image.
Inspired by this cognitive mode, we hypothesize that it is feasible and effective to recognize the facial expressions by a sequence of visual words from a global perspective.
Therefore, we propose the Visual Transformers with Feature Fusion (VTFF) for robust facial expression recognition in the wild.
Our method combines LBP features and CNN features to further enrich the representation of the visual words, referring to the hybrid feature extraction.
The reason we use LBP features is that it can catch the small movements of the faces and extract image texture information.
We design the attentional feature fusion (ASF) to adaptively integrate LBP features and CNN features.
The ASF aggregates both global and local relationships between two kinds of features, which can effectively improve the recognition performance.
We simply convert the fused feature maps into a sequence of visual words by flattening and projecting the features maps.
After obtaining these visual words, we then exploit the multi-layer Transformer encoder to boost the performance.
The global self-attention in the multi-layer Transformer encoder allows the network to model the contextual information of the representative visual words and focus on the most discriminative features.

Overall, the main contributions of our work can be summarized as follows:
\begin{enumerate}
  \item We propose the Visual Transformers with feature fusion for FER in the wild, which integrates LBP features and CNN features with the global-local attention and the global self-attention for improving expression recognition accuracy.
  \item We design a simple but effective feature fusion module named ASF to aggregate both global and local facial information. Moreover, the ASF guides the backbones to extract the required information while squeeze the useless information in an end-to-end manner.
  \item To the best of our knowledge, we are the first to apply Transformers for the FER. The global self-attention enables the whole network to learn the relationships between elements of visual feature sequences and ignore the information-deficient regions.
  \item Extensive experimental results on three publicly available FER in the wild datasets, i.e., RAF-DB, FERPlus and AffectNet, demonstrate that our VTFF achieves state-of-the-art expression recognition performance. Especially, we also conduct experiments on occlusion and variant pose subsets of these three datasets and cross-dataset evaluation on CK+ to show the promising generalization ability of our method.
\end{enumerate}

The rest of this article is organized as follows: Section \ref{relatedwork} briefly reviews related works for FER in the wild and provides a comprehensive review of recent advances in FER.
The core idea of our proposed VTFF is presented in Section \ref{method}.
Section \ref{sec:experiments} presents an extensive performance evaluation for the proposed method and state-of-the-art approaches.
We conclude our work in Section \ref{conclusion}.

\section{Related Work\label{relatedwork}}

\subsection{Facial Expression Recognition in the Wild}
Facial expression in the wild is significantly challenging when facing unexpected conditions mentioned in Section \ref{sec:introduction}.
Researchers start to shift their attention from hand-crafted features \cite{dalal2005histograms},\cite{shan2005robust},\cite{feng2005facial},\cite{buciu2004application},\cite{lee2014intra} to deep features \cite{he2016deep},\cite{huang2017densely} for accurately extracting discriminative features.
For example, Tang\cite{tang2013deep} utilized the CNNs for feature extraction and replaced the softmax layer with the linear SVMs, which gave significant gains and won the ICML 2013 Representation Learning Workshop's face expression recognition challenge.
Later on, Li et al. \cite{li2017reliable} proposed to enhance the discriminative power of deep features by a deep locality-preserving CNN (DLP-CNN) method.
These two approaches \cite{tang2013deep},\cite{li2017reliable} are remarkable, indicating the effectiveness of deep features for FER.
Meanwhile, region-based attention networks are especially suitable for FER in the wild, because they allow for salient face features to dynamically come to forefront when some occlusions or clutters occur in an image.
In\cite{wang2020region}, Wang et al. proposed region attention networks (RAN) to capture the importance of facial regions for occlusion and pose variant FER.
Likewise, Li\cite{li2018occlusion} et al. proposed a CNN with attention mechanism for Occlusion Aware FER, which focused on the most discriminative face regions.
Similar with \cite{wang2020region, li2018occlusion}, our method utilizes the global self-attention mechanism to recognize facial expressions from a global perspective, which emphasizes the discriminative regions and generalizes well in case of occlusions or variant poses.
Apart from these attribute-unrelated methods, some FER algorithms integrated demographic features (i.e., gender, race, age, etc.) for improving the expression recognition performance.
It is worth mentioning that Fan\cite{fan2020facial} firstly proposed a deeply-supervised attention network, which takes facial attributes into consideration.
In addition, Xu et al. \cite{xu2020investigating} proposed to use the attribute information as input to address bias and conducted a comparative study of the bias and fairness for FER.
To further suppress the uncertainties containing in the datasets, Wang et al. \cite{wang2020suppressing} proposed a self-cure Network (SCN) for FER in the wild, which took full advantage of attention mechanism to weight each training face sample.
These mainstream deep-learning based FER works focus on directly adopting the CNNs for extracting deep features, ignoring the carefully-designed features (such as LBP) for FER and the inter-relationships within deep features.

\subsection{Feature Fusion for FER}
Most research works only use deep features to distinguish facial expressions.
As we mentioned in Section \ref{sec:introduction}, there are various hand-crafted features, such as LBP\cite{moore2011local} and HOGs\cite{hu2008multi}, which have been well-developed for facial expression feature representation.
In addition, feature fusion can enrich the representative ability of the whole networks, which effectively improves the generalization ability and the recognition performance\cite{pong2014multi},\cite{lin2017feature},\cite{lin2017refinenet}.
Therefore, various feature fusion based FER methods have been proposed to further improve the performance of expression recognition.
Chen et al. \cite{chen2016facial} proposed to fuse dynamic textures, geometric features and acoustic features to tackle FER in the wild.
The dynamic texture descriptor of visual information is an extension of HOGs, named Histogram of Oriented Gradients from Three Orthogonal Planes (HOG-TOP).
In addition, Shao and Qian\cite{shao2019three} proposed a dual branch CNN to extracts LBP features and deep features parallel.
They utilized the concatenation operation to fuse these feature maps.
Likewise, Li et al. \cite{li2020attention} combined LBP features and CNN features with dense connection to improve expression recognition accuracy.
They did not well utilize complementary information and abandon redundant information.
Additionally, these feature fusion methods were mainly conducted on lab-controlled FER datasets, such as CK+ \cite{lucey2010extended}  and Oulu-CASIA \cite{zhao2011facial}.
Base on these, We propose the attentional selective fusion (ASF) for integrating the LBP features and the CNN features, which squeezes the useless information and generates correlated weight maps from both local and global perspectives.

\begin{figure*}[htbp] 
  \centering 
  \includegraphics[scale=0.7]{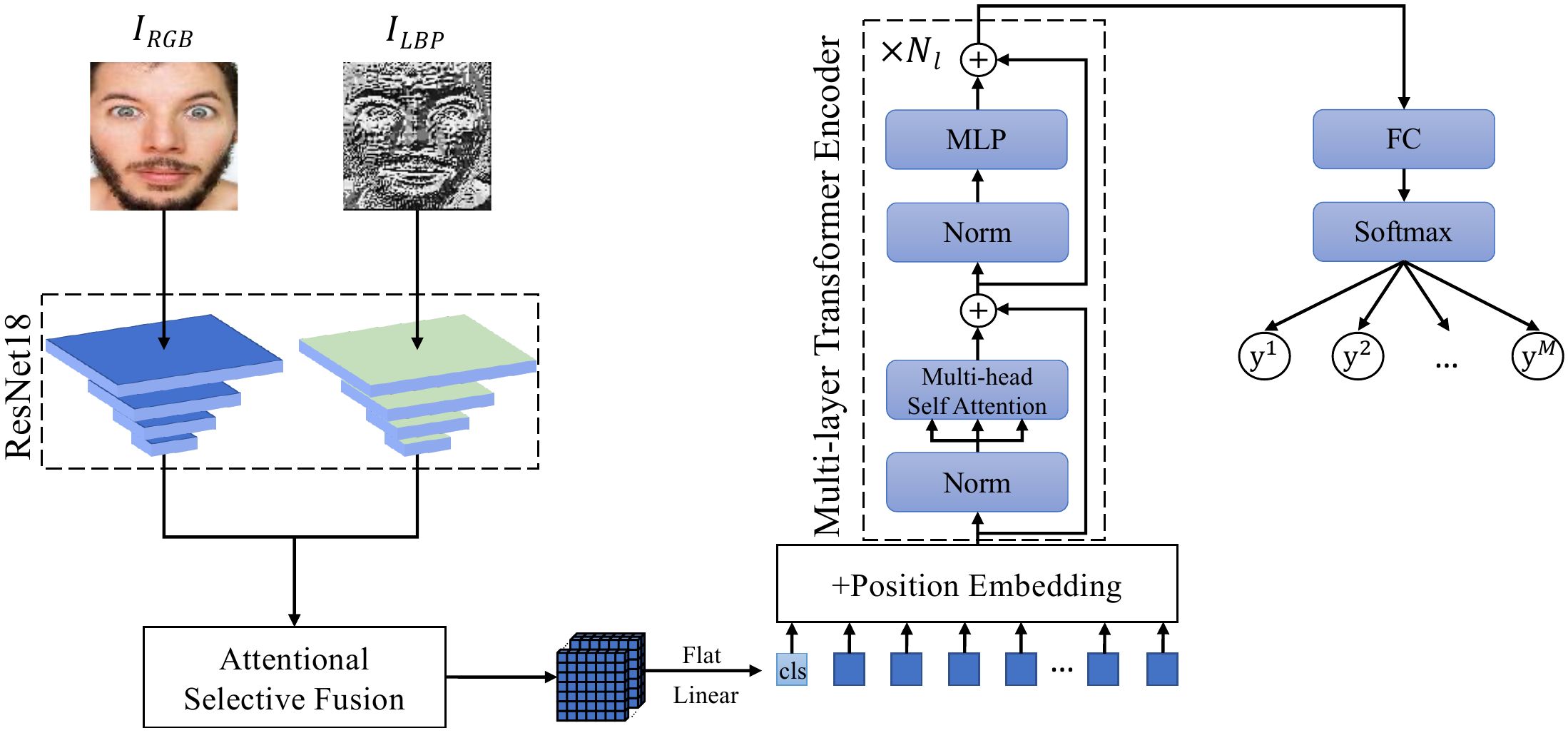} 
  \caption{An overview of our proposed VTFF. 
It can be divided into three parts, visual words extraction, relationship modeling and expression classification.
The pre-trained ResNet18 is used as the backbone to extract feature maps.
All the extracted features are fused by our attentional selective fusion to get representative visual words.
The input visual words are obtained by simply flattening the spatial dimensions of the feature maps and projecting to the specific dimension.
And then apply the multi-layer Transformer encoder to model the relationships between different visual features components.
The network finally calculates the expression probabilities by a simple softmax function.
  }
  \label{cvt}
\end{figure*}
\begin{figure}[ht]
  \centering 
  \includegraphics[scale=0.6]{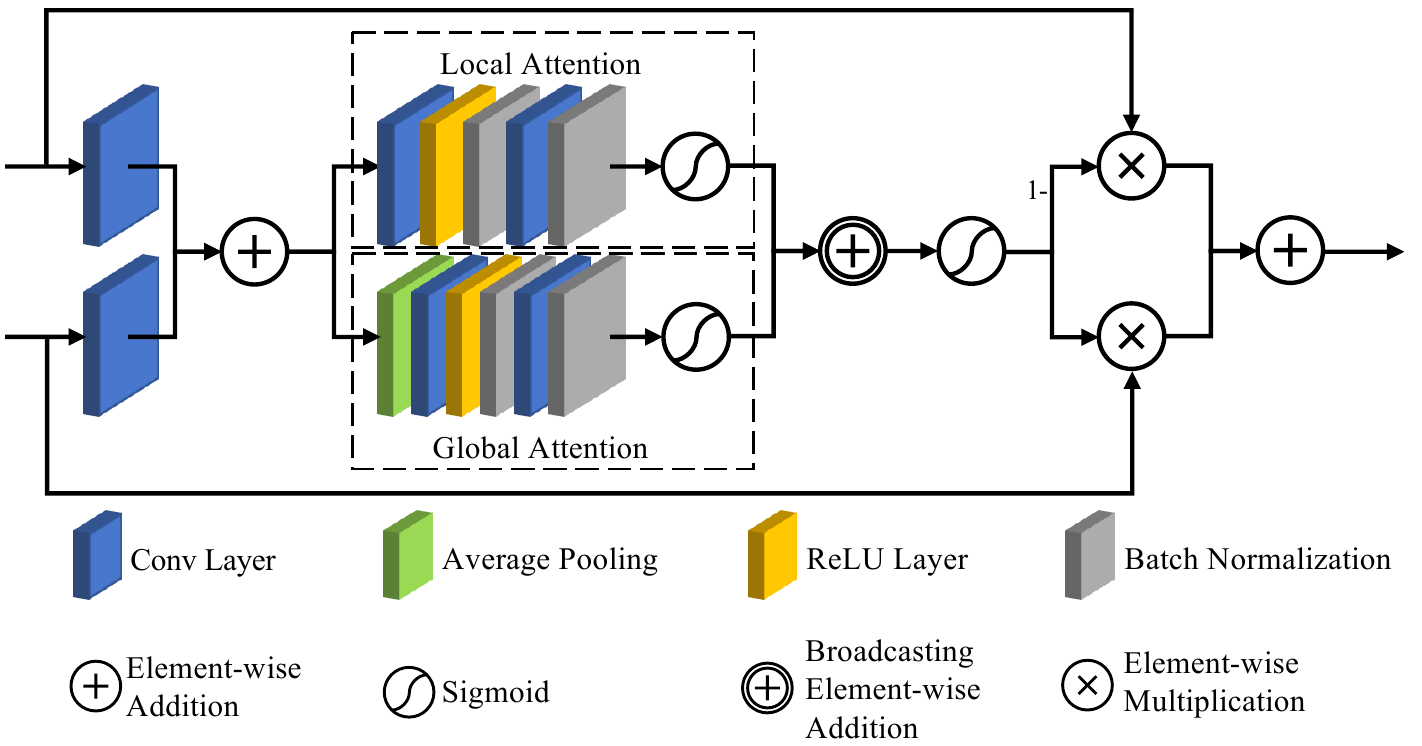} 
  \caption{The attentional selective fusion module.
  }
  \label{asf_fig}
\end{figure}

\subsection{Transformers in Computer Vision}
Apart from the CNNs-based feature representation technique, Transformers \cite{vaswani2017attention} have shown dominant performance in NLP.
Inspired by the success of Transformers, several researchers have tried to invest Transformers on computer vision tasks, such as object detection\cite{beal2020toward}, pose estimation\cite{yang2020transpose}, high-resolution image synthesis\cite{esser2020taming}, video instance segmentation\cite{wang2020end}, trajectory prediction\cite{bhat2020trajformer}, etc.
Transformer-based approaches have shown superior performance compared with CNN-based methods, when fully trained on large-scale datasets.
Vision Transformer (ViT)\cite{dosovitskiy2020image} was the first work to apply a vanilla Transformer to images with few modifications.
According to \cite{dosovitskiy2020image}, ViT yielded lower accuracy compared with ResNet when trained on ImageNet\cite{deng2009imagenet}.
ViT was firstly trained on large datasets, and then fine-tuned for downstream tasks, because Transformers need amounts of data to generalize well on computer vision tasks.
The feature pyramid structure in CNNs has also been applied with Transformers.
Wang et al. \cite{wang2021pyramid} proposed Pyramid Vision Transformer (PVT) for pixel-level dense prediction, which can work as the feature extraction backbone without convolutions.
In addition to these Transformers without convolution operation, several works \cite{chen2021visformer,wu2021cvt,liu2020convtransformer} proposed to blend convolutional layers into Transformers, which further improved the performance of pure Transformers.
Inspired by the vanilla Transformer and these brilliant Transformer-variants, we firstly propose to directly apply Transformers for FER.
As far as we know, no works attempt to capture the relationships among deep features for facial expression recognition.
We utilize Transformers to model long dependencies between input sequences by the global self-attention mechanism.
Such global self-attention enables the model to ignore the information-deficient regions and recognize the expressions from a global perspective in case of occlusions or variant poses for FER.

\section{Method\label{method}}


\subsection{Overview}\label{section:overview}
Fig. \ref{cvt} illustrates the overall diagram of our Visual Transformers with feature fusion for facial expression recognition.
Our VTFF is built upon on two pre-trained ResNet18\cite{he2016deep} networks, and consists of two crucial components: i) attentional selective fusion, ii) multi-Layer Transformer encoder.

For a given face image ${I_{RGB}}$ with the size of $H\times W\times 3$, we first get its LBP feature image with the size of $H\times W\times 1$ and concatenate it to a feature image ${I_{LBP}}$ with the size of $H\times W\times 3$.
The feature extraction backbones are composed of two ResNet18 networks: one is for the RGB image and the other is for its LBP feature image.
Particularly, we employ the first five stages of ResNet18 as the backbone to extract feature maps ${X_{LBP}}$ and ${X_{RGB}}$ with the size of $\frac{H}{R}\times \frac{W}{R}\times C_{f}$, where R is the downsampling rate of ResNet18, $C_{f}$ is the channel number of the output of the stage 5.
For simplicity, we denote $H_{d} = \frac{H}{R}$ and $W_{d} = \frac{W}{R}$.
In this paper, $R$ = 32 and $H$ = $W$ = 224.
We initialize the whole network weights by the pre-trained weights on MS-Celeb-1M face recognition dataset.
Without loss of generalization, the ASF is utilized to combine the features extracted from the RGB image and the features extracted from its LBP feature image, which will be introduced in Section \ref{section:asf} in detail.
The ASF module dynamically adjusts the weights of these features and guides the networks focus more on discriminative features that are vital for improving expression recognition.
The fusion weights of the ASF are generated via the global-local attention, which aggregates global and local context for further expression recognition.
The size of fused feature maps $ X_{fused}$ is also $H_{d}\times W_{d} \times C_{f}$.
Afterwards, we feed the flattened features to a linear projection and a learnable classification token is added.
We get embedded visual words with size of $(H_{d}W_{d} +1)\times C_{p}$, where $C_{p}$ is the channel of flattened features after projection.
We also add position embeddings to the embeddings to retain positional information, as \cite{vaswani2017attention} and \cite{dosovitskiy2020image} do.
The input embeddings are further fed to the Transformer encoder, which is composed of $N_{l}$ encoder blocks.
Finally, the probabilities of facial expressions are generated by a fully connected layer and the softmax function.

\subsection{Attentional Selective Fusion}\label{section:asf}
Our attentional selective fusion consists of global attention and local attention, as can be seen in Fig. \ref{asf_fig}, which can provide additional flexibility in fusing different types of information.
As mentioned in the Section \ref{section:overview}, given two feature maps ${X_{LBP}}, {X_{RGB}} \in \mathbb{R}^{H_{d}\times W_{d} \times C_{f}}$ extracted from the backbones, we first fuse the LBP features ${X_{LBP}}$ and the CNN features ${X_{RGB}}$ for capturing the subsequent information interaction:
\begin{equation}
  {U} = W_{L} {X_{LBP}} + W_{C} {X_{RGB}},
\end{equation}
where ${U}$ is the integrated feature maps after summation between ${X_{LBP}}$ and ${X_{RGB}}$, and $+$ denotes element-wise summation.
$W_{L}$ and $W_{C}$ are the weights for initial integration and simply implemented by two $ 1\times 1$ convolutions.

To perform both global and local selective fusion, we then choose global average pooling and the pixel-wise convolution as global context and local context aggregator, respectively.
The global context progressively squeezes each feature map of size $H_{d} \times W_{d}$ into a scalar,
and exploits the inter-channel relationship of features.
Different from the global context, the local context preserves and highlights the subtle details of the input features, which is complementary to the global context.
Aggregating local and global contexts facilitates the network to benefit from different types of features and recognize ambiguous facial expressions more accurately.
The global context and local context are computed as follows:
\begin{align}
  \label{eq1}
  {G(U)} &= \sigma (\mathbb{B} \mathbb{N}({ Conv^{2}_{G}}(\delta (\mathbb{B} \mathbb{N} ({ Conv^{1}_{G}}(\mathbb{A} \mathbb{P} ({U}))))))),\\
  \label{eq2}
  {L(U)} &= \sigma (\mathbb{B} \mathbb{N}({ Conv^{2}_{L}}(\delta (\mathbb{B} \mathbb{N} ({ Conv^{1}_{L}}({U})))))).
\end{align}
As Eq. \eqref{eq1} shows, the global adaptive average pooling $\mathbb{A} \mathbb{P}$ is first applied to squeeze each feature map of ${U} \in \mathbb{R}^{H_{d}\times W_{d}\times C_{f}}$ into a scalar on the spatial dimension,
which is carried out using Eq. \eqref{gap}.
\begin{equation}
  \label{gap}
  \mathbb{A} \mathbb{P} ({U}) = \frac{1}{H_{d}W_{d}}\sum_{i = 1}^{H_{d}}\sum_{i = 1}^{W_{d}}x_{c}(i,j), c=1,2,\dots,C_{f}, 
\end{equation} 
where $H_{d}$ and $W_{d}$ are the height and width of the input feature map ${U}$, and $C_{f}$ is the number of channels of ${U}$.
$\mathbb{B} \mathbb{N} $ is the Batch Normalization. $\delta$ denotes the ReLU function, and $\sigma$ denotes the Sigmoid function.
The kernel sizes of ${ Conv^{1}_{G}}$ and ${ Conv^{2}_{G}}$ are $ \frac{C_{f}}{r} \times C_{f} \times 1\times 1$, $ C_{f} \times  \frac{C_{f}}{r} \times 1\times 1 $.
And the kernel sizes of ${ Conv^{1}_{L}}$, ${ Conv^{2}_{L}}$ are $\frac{C_{f}}{r}\times C_{f} \times 1\times 1  $, $ 1\times \frac{C_{f}}{r} \times  1\times 1$.
Although $Conv^{1}_{G}$ and $Conv^{1}_{L}$ have the same kernel size, their functions on the inputs are significantly different, especially when coupled with $Conv^{2}_{G}$ and $Conv^{2}_{L}$.
Specifically, $Conv^{1}_{G}$ with the size of $\frac{C_{f}}{r}\times C_{f} \times 1\times 1 $ squeezes its input with the reduction ratio $r$. We set $r$ to 8 in this paper.
And $Conv^{2}_{G}$ increases the channel dimensionality of its input and later results in the global fusion weights $G(U) \in \mathbb{R}^ {1 \times 1 \times C_{f}}$.
Therefore, $Conv^{1}_{G}$ and $Conv^{2}_{G}$ are applied to capture the global relationships of these features along the channel dimension.
Meanwhile, $Conv^{1}_{L}$ squeezes ${U} \in \mathbb{R}^{H_{d}\times W_{d}\times C_{f}}$ into the shape of $H_{d}\times W_{d}\times \frac{C_{f}}{r}$.
And $Conv^{2}_{L}$ also squeezes the input along the channel dimension while keeping the spatial dimensionality of its input constant.
The shape of the $Conv^{2}_{L}$'s output is $H_{d} \times W_{d} \times 1$.
In other words, $Conv^{1}_{L}$ and $Conv^{2}_{L}$ progressively reduce the channel dimensionality of features maps to focus on the discriminative information scattered locally.

Given the global fusion weights $G(U)$ and the local fusion weights $L(U)$, the refined global-local attention weights can be obtained by Eq. \eqref{global-local}.
\begin{equation}
  \label{global-local}
   GL(U) = G(U)\oplus L(U),
\end{equation}
where $\oplus$ represents the broadcasting addition. Then, the fused feature map $ X_{fused}$ is calculated by as follows:
\begin{equation}
  \label{fuse}
  \begin{aligned}
     X_{fused} = &{X}_{ LBP}\otimes \sigma ( GL(U)) +\\
     & {X}_{RGB}\otimes \sigma (1- GL(U)),
  \end{aligned} 
\end{equation}
where $\otimes$ is the element-wise multiplication.

\subsection{Multi-Layer Transformer Encoder}
The fused 2D feature map $ X_{fused}$ need to be flattened into a 1D visual embedding sequence, and further can be fed for the multi-layer Transformer encoder as input.
Therefore, we reshape $X_{fused}\in \mathbb{R}^{H_{d}\times W_{d}\times C_{f}}$ into a flattened sequence and feed it to a linear projection to get $X_{f} \in \mathbb{R}^{H_{d}W_{d}\times C_{p}}$, where $H_{d}W_{d}$ is the sequence length and $C_{f}$, $C_{p}$ are set to 512 and 768 respectively.
As in \cite{dosovitskiy2020image}, a classification token [cls] is appended at the beginning of the input sequence $X_{f}$.
The learnable state of the [cls] token at the output of the Transformer encoder is utilized to represent the whole feature sequence, which serves for the final prediction.
To incorporate the positional information in the multi-layer Transformer encoder, the 1D learnable positional embeddings are added to the feature embeddings:
\begin{equation}
    Z^{0} = [x_{cls};x_{f}^{1};x_{f}^{2};x_{f}^{3};\dots;x_{f}^{H_{d}W_{d}}] + PE(H_{d}W_{d}+1;C_{p}),
\end{equation}
where $PE(H_{d}W_{d}+1;C_{p}) \in \mathbb{R}^{(H_{d}W_{d}+1)\times C_{p}}$ learns the embeddings for each position index, [cls] token included, and $Z^{0}$ represents the resulting position-aware feature sequence.

To model the complex interactions among all elements of the facial feature embeddings, we input $Z^{0}$ to the standard multi-Layer Transformer encoder.
The Transformer encoder calculates the weights of embeddings $Z^{0}$ through multi-head self-attention (MHSA).
This is done by learnable queries $Q$, keys $K$, and values $V$.
We compute the single-head global self-attention(SHSA) using Eq. \eqref{softmax_attention}.
Details of SHSA in the first layer can be formulated as follows:
\begin{equation}
  \label{softmax_attention}
  \begin{aligned}
    head_{j} &= Attention(Q_{j},K_{j},V_{j})\\
    &= softmax(\frac{Q_{j}K_{j}^{T}}{\sqrt{d}})V_{j}\\
    &= softmax(\frac{Z^{0}W_{j}^{Q}(Z^{0}W_{j}^{K})^{T}}{\sqrt{d}})Z^{0}W_{j}^{V},
  \end{aligned}
\end{equation}
where $Q_{j} = Z^{0}W_{j}^{Q}$, $K_{j} = Z^{0}W_{j}^{K}$, $V_{j}=Z^{0}W_{j}^{V}$ and $W_{j}^{Q}\in \mathbb{R}^{C_{p}\times d} $, $W_{j}^{K}\in \mathbb{R}^{C_{p}\times d}$, $W_{j}^{V}\in \mathbb{R}^{C_{p}\times d}$ are the parameters of these linear projections.
Specifically, multi queries, keys and values project $Z^{0}$ into $N_{h}$ different representation subspaces.
Multi-head self-attention (MHSA) can be described as:
\begin{align}
  \label{mhsa}
  { MHSA(Z^{0})} = concat(head_{1},\dots,head_{N_{h}})W^{O},
\end{align}
where $N_{h}$ is the number of different heads, and $concat$ denotes the concatenation operation.
$W^{O} \in \mathbb{R}^{h_{1}\times d}$ are the parameters of a linear projection, where the dimension of each head $d$ is equal to $\frac{C_{p}}{N_{h}}$ and $h_{1}$ is the hidden size of the first layer.
Each Transformer encoder consists of $N_{l}$ layers of MHSA blocks.
Formally, the standard multi-layer Transformer encoder computes forwardly for $i=1,\dots,N_{l}$ layers:
\begin{align}
  \hat{Z}^{i} &= { MHSA(LN}(Z^{i-1})) + Z^{i-1}\\
  Z^{i} &= { MLP(LN}(\hat{Z}^{i})) + \hat{Z}^{i}
\end{align}
where $\hat{Z}^{i}$ and $Z^{i}$ are intermediate output and final output at layer $i$.
The $MLP$ consists of two position-wise feed-forward layers and a GELU non-linearity activation function.
The $LN$ denotes the layer normalization, which is applied before every attention block and the $MLP$.
The hidden dimension of the $MLP$ is set to 3,072 in this paper.
The output $Z^{N_{l}}_{0}$ of the layer $N$ is also normalized by the $LN$.
It is notable that we just apply the fully-connected layer for the final [cls] token, which is further for calculating the expression probability scores.
Mathematically, the probability scores are generated as follows:
\begin{align}
  Y &= LN(Z^{N_{l}}_{0}),\\
  y^{i} &= \frac{e^{\theta_{i}^{T}Y}}{\sum_{i=1}^{M}e^{\theta_{i}^{T}Y}},
\end{align}
where $Z^{N_{l}}_{0}$ is the first [cls] token of the whole sequence $Z^N_{l}$ and $Y$ is the output of the Multi-layer Transformer encoder.
$\theta$ represents the parameters of the fully-connected layer and $\theta_{i}$ is the $i$-th column of $\theta$.
The number of expression classes is $M$.
$y^{i}$ is the probability score of the $i$-th facial expression, and the final predicted expressions can be easily obtained by $\mathop{\arg\max}$ function during inference.

\section{Experiments}\label{sec:experiments}
In order to demonstrate the effectiveness of our proposed method, we carry out extensive experiments on three in-the-wild FER datasets (i.e., RAF-DB, FERPlus and AffectNet) and cross-dataset evaluation on CK+.
In this section, we first introduce the FER datasets used in our experiments and implementation details.
Then, the proposed method is compared with several state-of-the-art approaches.
Subsequently, the impact of each component of the proposed VTFF model is investigated with experiments on these datasets.
\subsection{Datasets}
We evaluate our approach on three frequently used facial expression datasets (RAF-DB, FERPlus and AffectNet).
These datasets are all collected in the wild, which may suffer from different illuminations and occlusions.
To demonstrate the effectiveness of our method when handling occlusion and variant pose issues in real-word conditions, we also conduct experiments on Occlusion-RAF-DB, Pose-RAF-DB, Occlusion-FERPlus, Pose-FERPlus, Occlusion-AffectNet, Pose-AffectNet, which are subsets of RAF-DB, FERPlus and AffectNet, respectively.
The cross-dataset evaluation experiments are also conducted on CK+ to verify the superior generalization ability of our method.
The details of the datasets used in the experiments are introduced as follows.

\textbf{RAF-DB} contains 29,672 real-world facial images collected from Flickr.
The whole images of RAF-DB are labeled by 315 well-trained annotators and each image is labeled by about 40 independent annotators.
RAF-DB contains two different subsets: single-label subset and multi-label subset.
In our experiments, we only use single-label subset, including seven basic emotions (neutral, happy, surprise, sad, angry, disgust, fear).
The images from the single-label subset are split into 12,271 training samples and 3,068 testing samples.
The expressions in both training images and test images have imbalanced distribution.
The overall sample accuracy is used for performance measurement.

\textbf{FERPlus} is extended from the original FER2013 dataset, which was for the ICML 2013 Challenges in Representation Learning.
FERPlus contains 28,709 training images and 3,589 test images, all of which are collected by the Google search engine.
The original size of the images in FERPlus is 48 $\times$ 48.
Each image in FERPlus is annotated by 10 annotators and FERPlus provides better quality labels than the original FER2013 labels.
Apart from seven basic emotions as RAF-DB, the contempt category is included in the labels.
We mainly report overall sample accuracy under the supervision of majority voting for performance measurement.

\textbf{AffectNet} is the largest facial expression datasets with more than 1,000,000 facial images collected from the Internet.
AffectNet provides both discrete categorical and continuous dimensional (i.e., valence and arousal) annotations.
It should be noted that AffectNet has imbalanced training, validation and test sets, of which 450,000 images have been annotated manually.
In our experiment, we utilize images annotated with eight basic expressions as FERPlus, 287,652 images for training and 4,000 images for testing.
Since the test set is not available to the public, we mainly report mean class accuracy on the validation set for performance measurement and fair comparison with other methods.

\textbf{Occlusion and Pose Variant Datasets} (i.e., Occlusion-RAF-DB, Pose-RAF-DB, Occlusion-FERPlus, Pose-FERPlus, Occlusion-AffectNet, Pose-AffectNet) are occlusion and pose subsets of RAF-DB, FERPlus, AffectNet and manually collected by \cite{wang2020region}.
There are various occlusion types occurring in samples of these datasets, such as wearing mask, wearing glasses and objects in upper/bottom face.
In addition, variant pose issues can be divided into two categories, poses larger than 30 degrees and poses larger than 45 degrees.
The sample distribution can be found in \cite{wang2020region}.
We report overall sample accuracy or mean class accuracy according to corresponding original datasets.

\textbf{CK+} is the extended Cohn-Kandade(CK) dataset for facial action unit and expression recognition, collected under a lab-collected environment.
The original data of CK+ has 593 video sequences from 123 subjects, 327 of which are annotated with seven basic emotions and contempt.
Each of the video sequences consists of images from onset (the first frame) to peak expression (the last frame).
We follow previous work to utilize the first frame of a video sequence as a neutral sample and the last frame with the target expression.
In total, we obtain 618 images with seven emotions and 654 images with eight emotions for testing.
We use the overall sample accuracy to evaluate the generalization capacity of our method.

\begin{table*}[htbp]
  \centering
  \caption{Comparison with state-of-the-art methods on RAF-DB. The best results are in bold.}
  \begin{tabular}{lccccccccc}
        \toprule
    method  &Year   & Angry & Disgust & Fear  & Happy & Sad   & Surprise & Neutral & Accuracy \\
         \midrule
  VGG\cite{li2017reliable}       & 2018  & 66.05 & 25.00   & 37.84 & 73.08 & 51.46 & 53.49    & 47.21   & 69.34    \\
  baseDCNN\cite{li2017reliable}   & 2018  & 70.99 & 52.50   & 50.00 & 92.91 & 77.82 & 79.64    & 83.09   & 82.66    \\
  Center Loss\cite{li2017reliable} & 2018  & 68.52 & 53.13   & 54.05 & 93.08 & 78.45 & 79.63    & 83.24   & 82.86    \\
  DLP-CNN\cite{li2017reliable}     & 2018  & 71.60 & 52.15   & 62.16 & 92.83 & 80.13 & 81.16    & 80.29   & 82.74    \\
  FSN\cite{zhao2018feature}       &  2018   & 72.80 & 46.90   & 56.80 & 90.50 & 81.60 & 81.80    & 76.90   & 81.14    \\
  gACNN\cite{li2018occlusion}   &  2018    &    -   &   -      &   -    &    -   &   -    &    -      &    -     & 85.07    \\
  RAN\cite{wang2020region}     &   2020   &  -     &    -     &    -   &   -    &    -   &    -      &   -      & 86.90    \\
  SCN\cite{wang2020suppressing}     & 2020     &   -    &     -    &    -   &   -    &   -    &    -      &    -     & 87.03   \\
  DSAN-VGG-RACE\cite{fan2020facial}& 2020& 82.71 & 56.25   & 58.11 & 94.01 & 83.89 & \textbf{89.06}    & 80.00   & 85.37    \\
  SPWFA-SE\cite{li2020facial}&	2020    &80.00   &	59.00	& 59.00	 & 93.00&84.00	&88.00	   &86.00    	&86.31\\
  \midrule
  Ours    & 2021   &  \textbf{85.80} &  \textbf{68.12} & \textbf{64.86}  &  \textbf{94.09}  & \textbf{87.24}  & 85.41 & \textbf{87.50} & \textbf{88.14}  \\
  \bottomrule
  \end{tabular}
  \label{raf_results}
\end{table*}

\begin{table}[htbp]
  \caption{Comparison with state-of-the-art methods \\on FERPlus and AffectNet. The best results are in bold.}
  \centering  
  \subtable[Results on FERPlus.]{    
    \label{ferp_results}      
    \begin{tabular}{p{2.5cm}cc}
      \toprule
      Method  & Year & Accuracy \\
      \midrule
      CSLD\cite{barsoum2016training} & 2016 & 83.85    \\
      ResNet+VGG\cite{huang2017combining}     & 2017 & 87.4    \\
      SHCNN\cite{miao2019recognizing}  & 2019 & 86.54    \\
      LDR\cite{fan2020learning}  & 2020 & 87.6    \\
      RAN$^{\diamond }$\cite{wang2020region}    & 2020 & 88.55   \\
      RAN\cite{wang2020region}     & 2020 & 87.85   \\
      SCN\cite{wang2020suppressing}     & 2020 & 88.01   \\
      \midrule
      Ours    & 2021 & \textbf{88.81}   \\
      \bottomrule
    \end{tabular}
  }  
  \qquad  
  \subtable[Results on AffectNet.]{  
    \label{affectnet_results}      
    \begin{tabular}{p{2.5cm}cc}
      \toprule
      Method  & Year & Accuracy \\
      \midrule
      IPA2LT\cite{zeng2018facial} & 2018 & 55.11   \\
      gACNN\cite{li2018occlusion}   & 2018 & 58.78   \\
      SPWFA-SE\cite{li2020facial} & 2020 & 59.23    \\
      RAN$^{\dagger }$  \cite{wang2020region}    & 2020 & 52.97   \\
      RAN\cite{wang2020region}    & 2020 & 59.50    \\
      SCN\cite{wang2020suppressing}     & 2020 & 60.23    \\
      \midrule
      Ours$^{\dagger }$  &2021 & 56.13\\
      Ours & 2021 & \textbf{61.85}   \\
      \bottomrule
    \end{tabular}
    }
  \end{table}

\subsection{Implementation Details}
In our experiments, the face images are detected by MTCNN\cite{zhang2016joint} and further resized to the size of 224 $\times$ 224.
For fair comparison with previous state-of-the-art methods, we use the same backbone ResNet18 pre-trained on the MS-Celeb-1M face recognition dataset.
The facial features are extracted from the last convolutional stage of ResNet18.
The learning rate of our methods is initialized as 0.005. We use a linear learning rate warmup of 1,000 steps and cosine learning rate decay.
The Adam optimizer \cite{kingma2014adam} is used to optimize the whole networks with a batch size of 32 and train the model for 20,000 steps on RAF and FERPlus, 40,000 steps on AffectNet, respectively.
The standard cross-entropy loss is utilized to supervise the model to generalize well for expression recognition.
We implement our method with Pytorch\cite{paszke2019pytorch} toolbox and conduct all the experiments on a single NVIDIA GTX 1080Ti GPU card.

\subsection{Comparison with State-of-the-art Methods}

The proposed VTFF model is compared with several state-of-the-art methods on RAF-DB, FERPlus and AffectNet.
We achieve new state-of-the-art results on these datasets to our knowledge.
The better performance demonstrates the superiority of our proposed method.

\textbf{Results on RAF-DB:} Comparison with other state-of-the-art methods can be found in Table \ref{raf_results}.
The methods in \cite{li2017reliable} presented their performance using mean accuracy. For fair comparison, we convert their results to accuracy, following \cite{li2020facial}.
Since gACNN\cite{li2018occlusion}, RAN\cite{wang2020region} and SCN\cite{wang2020suppressing} did not report the specific expression recognition accuracy or the confusion matrices, the corresponding places are marked with '-' in Table \ref{raf_results}.
Although SPWFA-SE\cite{li2020facial} also did not report specific expression recognition accuracy, it provided the confusion matrix on RAF-DB.
Therefore, we borrow the accuracy results from its confusion matrix for comparison.
Overall, our proposed method achieves 88.14\% on RAF-DB.
As shown in Table \ref{raf_results}, our method achieves all the best results among all methods except for the surprise category.
In detail, our VTFF has obtained gains of 18.8\% and 1.11\% over VGG and SCN, which are the baseline method and the previous state-of-the-art method, respectively.
DSAN-VGG-RACE integrated deeply-supervised blocks and attention blocks with race labels, which were additional data compared with our purely used expression labels.
Since RAF-DB also has extremely imbalanced distribution, the slight performance decline of the surprise category is reasonable and acceptable.
The accuracy of the disgust expression recognition of our method has recorded an increase of 9.12\% compared with the previous best result of \cite{li2020facial}, which demonstrates the effectiveness and the superiority of our method in feature learning.

\textbf{Results on FERPlus:} Table \ref{ferp_results} presents the comparison results on FERPlus.
We compare our models with CNN-based methods including CSLD \cite{barsoum2016training}, ResNet+VGG \cite{huang2017combining}, SHCNN \cite{miao2019recognizing}, LDR \cite{fan2020learning}, as well as two recent state-of-the-art methods(RAN \cite{wang2020region}, SCN \cite{wang2020suppressing}).
As we can see in Table \ref{ferp_results}, our proposed VTFF achieves 88.81\% on FERPlus.
Under the same experiment settings, total improvements of our VTFF on FERPlus are 0.96\% and 0.80\% when compared with RAN and SCN.
Especially, RAN$^{\diamond }$ means the improved RAN with extra face alignment.
Although face alignment is crucial for face recognition and facial expression recognition, it is a preprocessing step and CNN-based methods tend to recognize expression in an end-to-end manner.
Even without face alignment in our experiment settings, our VTFF also achieves much better results over RAN$^{\diamond }$.

\begin{table*}[!h]
  \caption{Comparison with lately published methods on AffectNet-7. 
  It is noted that the accuracy \\results are borrowed from their confusion matrices.
  The best results are in bold.}
  \centering
  \setlength{\tabcolsep}{2.5mm}{
  \begin{tabular}{lccccccccc}
        \toprule
    method  &Year   & Angry & Disgust & Fear  & Happy & Sad   & Surprise & Neutral & Accuracy \\
         \midrule
  MFMP+\cite{happy2021expression} &  2021   & 55.00 & 46.00   & 53.00 & 88.00 & 55.00 & 55.00    & 64.00     & 58.86    \\
  IDFL\cite{li2021learning} &  2021   & 31.00 & 65.00   & 49.00 & 95.00 & 59.00 & 43.00    & 73.00   & 59.20    \\
  WSFER\cite{zhang2021weakly} &  2021   & 58.54 & 30.28   & 50.42 & 88.01 & \textbf{67.56} & 51.90    & 73.55    & 60.04    \\
  T21DST\cite{xie2021triplet} &  2021   & 18.00 & 40.00   & 53.00 & \textbf{96.00} & 62.00 & 62.00    & \textbf{79.00}   & 60.12    \\
  SDW\cite{hayale2021deep}       &  2021   & 53.00 & \textbf{56.00}   & 61.00 & 86.00 & 58.00 & 53.00    & 59.00    & 61.11    \\
  ReCNN\cite{xia2021relation} &  2021   & 59.00 & 54.40   & \textbf{65.60} & 87.60 & 59.40 & 60.00    & 62.40    & 64.06    \\
  \midrule
  Ours    & 2021   &  \textbf{61.20} &  53.00 & 60.40  &  88.40  & 60.80  & \textbf{64.80} & 65.00  & \textbf{64.80}  \\
  \bottomrule
  \end{tabular}
  \label{aff_new}
  }
\end{table*}


\begin{table}[t]
  \caption{Comparison with other methods \\on Occlusion and Pose Variant Datasets.}
  \centering  
  \subtable[Results on Occlusion-RAF-DB, Pose-RAF-DB.]{    
    \label{subraf_results}      
    \begin{tabular}{p{2cm}ccc}
      \toprule
      Method  & Occlusion & Pose(30) & Pose(45)\\
      \midrule
      Baseline\cite{wang2020region} & 80.19 &  84.04& 83.15  \\
      RAN\cite{wang2020region}   & 82.72 & 86.74&85.20   \\
      \midrule
      Ours  & 83.95 & 87.97&  88.35 \\
      \bottomrule
    \end{tabular}
    }
  \qquad  
  \subtable[Results on Occlusion-FERPlus, Pose-FERPlus.]{  
    \label{subferp_results}      
    \begin{tabular}{p{2cm}ccc}
      \toprule
      Method  & Occlusion & Pose(30) & Pose(45)\\
      \midrule
      Baseline\cite{wang2020region} & 73.33 &  78.11& 75.50  \\
      RAN\cite{wang2020region}   & 83.63 & 82.23&80.40   \\
      \midrule
      Ours  & 84.79 & 88.29&  87.20 \\
      \bottomrule
    \end{tabular}
    }
    \qquad  
    \subtable[Results on Occlusion-AffectNet, Pose-AffectNet.]{  
      \label{subaffectnet_results}      
      \begin{tabular}{p{2cm}ccc}
        \toprule
        Method  & Occlusion & Pose(30) & Pose(45)\\
        \midrule
        Baseline\cite{wang2020region} & 49.48 &  50.10& 48.50  \\
        RAN\cite{wang2020region}   & 58.50& 53.90& 53.19  \\
        \midrule
        Ours  & 62.98 &60.61 & 61.00  \\
        \bottomrule
      \end{tabular}
      }
    \label{fer_results}
    \end{table}

\textbf{Results on AffectNet:} We compare our method with several methods on AffectNet and report the results in Table \ref{affectnet_results}.
We obtain 61.85\% with oversampling on AffectNet, without bells and whistles.
IPA2LT \cite{zeng2018facial}, gACNN \cite{li2018occlusion} and SPWFA-SE \cite{li2020facial} are trained for seven classes on AffectNet without the contempt category. 
As mentioned above, AffectNet has imbalanced distribution.
SPWFA-SE utilizes focal loss function to handle the imbalance problem.
To deal with the imbalance issue, as RAN \cite{wang2020region} and SCN \cite{wang2020suppressing} do, we adopt the oversampling strategy in our experiments.
Especially, method marked with $^{\dagger}$  represents it is trained without the oversampling strategy.
Our method without using oversampling exceeds RAN$^{\dagger}$ by 3.16\%.
In addition, the oversampling technique enhances RAN$^{\dagger}$ as well as our method by 6.53\% and 5.72\% respectively, which demonstrates the oversampling technique  greatly eliminates the negative effect of imbalanced class distribution.
Our VTFF with only AffectNet for training outperforms SCN by 1.62\%, which applied extra dataset WebEmotion for pre-training and then fine-tuned SCN on AffectNet.
The improvements of our VTFF over previous methods suggest that the VTFF indeed has better generalization ability even on large-scale expression recognition datasets like AffectNet.

\begin{table}[t]
  \centering
  \caption{Cross-dataset evaluation results on CK+.}
  \begin{tabular}{p{2cm}ccc}
    \toprule
    Method  & Train & Test & Accuracy\\
    \midrule
    gACNN\cite{li2018occlusion} & RAF-DB &  CK+& 81.07  \\
    SPWFA-SE\cite{li2020facial}   & RAF-DB& CK+& 81.72  \\
    SPWFA-SE\cite{li2020facial}   & AffectNet-7& CK+ & 85.44  \\
    \midrule
    Ours  & RAF-DB & CK+ & 81.88  \\
    Ours  & FERPlus & CK+ & 83.79  \\
    Ours  & AffectNet-8 & CK+ & 86.24  \\
    \bottomrule
  \end{tabular}
  \label{ck_results}
\end{table}

To further investigate our proposed VTFF on AffectNet, we compare the performance of VTFF with the lately published approaches including \cite{happy2021expression, li2021learning,zhang2021weakly,xie2021triplet,hayale2021deep,xia2021relation}.
To make fair comparison, we also train our model excluding the contempt category.
Table \ref{aff_new} illustrates the detail comparison results on AffectNet-7.
It can be seen that our proposed method achieves superior performance in terms of mean class accuracy compared with other methods with a 5.94\% to 0.74\% improvement.
Additionally, the proposed method also achieves the highest accuracies for the two facial expressions (Angry and Surprise) among these methods, which are 61.2\% and 64.80\% respectively.
Xia et al. \cite{xia2021relation} propose the Relation Convolutional Neural Network to focus on the most discriminative facial regions, which achieves an accuracy of 64.6\% on average and the highest accuracy in Fear (64.6\%).
The IDFL \cite{li2021learning} and the T21DST \cite{xie2021triplet} achieve higher accuracies in Neural (79.00\%, 73.00\%) and Happy (96.00\%, 95.00\%), but their performances in Angry (31.00\%, 18.00\%) and Fear (49.00\%, 53.00\%) are far from satisfactory. This is because they do not take the imbalanced class distribution into consideration.
Different from \cite{li2021learning, xie2021triplet}, Hayale et al. \cite{hayale2021deep} propose to set different penalization weights for classes based on the proportions in the training set, achieving the highest accuracy on Disgust (56.00\%).
Zhang et al.\cite{zhang2021weakly} leverage noisy data to boost the performance of FER and obtain the best result in Sad (67.56\%).
We can conclude from Table \ref{aff_new} that our proposed VTFF achieves competitive results and even outperforms the lately published methods, which shows the effectiveness of our VTFF.

\begin{table*}[t]
  \caption{Ablation study w.r.t. LBP features, ASF and MTE, performed on RAF-DB, FERPlus and AffectNet.\\
  It is noted that we set $N_{l} = 4$ and $N_{h} = 8$ to explore the effects of other modules and conduct a fair ablation study.\\ We run the experiment three times over multiple random seeds, and then compute the mean and standard deviation of the recognition accuracy.}\label{ablation_study}
  \centering
  \setlength{\tabcolsep}{5mm}{
  \begin{tabular}{c|cccccc}
    \toprule
    Setting & LBP & ASF & MTE & RAF-DB & FERPlus & AffectNet\\
    \midrule
    {\textbf{a}} &{{\XSolidBrush}}&{{\XSolidBrush}}&{{\XSolidBrush}}&{{86.37 $\pm$ 0.35 }}&{{86.70 $\pm$ 0.24 }}&{{58.45 $\pm$ 0.10}}\\
    {\textbf{b}} &{{\CheckmarkBold}}&{{\XSolidBrush}}&{{\XSolidBrush}}&{{86.53 $\pm$ 0.27}}&{{87.17 $\pm$ 0.30}}&{{58.65 $\pm$ 0.14}}\\
    {\textbf{c}} &{{\CheckmarkBold}}&{{\CheckmarkBold}}&{{\XSolidBrush}}&{{87.14 $\pm$ 0.39}}&{{87.55 $\pm$ 0.18}}&{{59.88 $\pm$ 0.21}}\\
    {\textbf{d}}      &{{\XSolidBrush}} &{{\XSolidBrush}}  &{{\CheckmarkBold}} &{{87.60 $\pm$ 0.12}}&{{87.65 $\pm$ 0.15}}&{{60.50 $\pm$ 0.09}}\\
    {\textbf{e}} &{{\CheckmarkBold}}&{{\XSolidBrush}}&{{\CheckmarkBold}}  &{{87.83 $\pm$ 0.04}}&{{88.15 $\pm$ 0.23}}&{{60.90 $\pm$ 0.12}} \\
    \midrule
    {\textbf{f}} &{{\CheckmarkBold}}&{{\CheckmarkBold}}&{{\CheckmarkBold}}&{{88.19 $\pm$ 0.21}}&{{88.70 $\pm$ 0.17}}&{{61.52 $\pm$ 0.07}} \\
    \bottomrule
  \end{tabular}
  }
\end{table*}

\textbf{Results on Occlusion and Pose Variant Datasets:} To examine our method in case of occlusion and variant pose in real scenarios, we also conduct several experiments on Occlusion-RAF-DB, Pose-RAF-DB, Occlusion-FERPlus, Pose-FERPlus, Occlusion-AffectNet and Pose-AffectNet.
Table \ref{fer_results} shows the accuracy of the experimental results under corresponding subsets.
RAN \cite{wang2020region} proposed to divide a face image into subregions and introduced a region biased loss for capturing the importance of different regions for occlusion and pose variant expression recognition.
Although our VTFF is not specifically designed for occlusion and variant pose FER issues, our method outperforms RAN with a large margin in each case, which shows the superiority of our method.
Specifically, our method exceeds RAN by 1.23\%, 1.16\% and 4.48\% on Occlusion-RAF-DB, Occlusion-FERPlus and Occlusion-AffectNet.
Our method also outperforms RAN on Pose-RAF-DB, Pose-FERPlus and Pose-AffectNet.
The gains are 1.23\%, 6.06\% and 6.71\% with pose larger than 30 degrees.
On Pose-FERPlus, Pose-AffectNet and Pose-RAF-DB with pose larger than 45 degrees, our method significantly outperforms RAN with the gains of 3.15\%, 6.8\% and 7.81\%, respectively.
Overall, these results reliably verify the effectiveness of our method on occlusion and variant pose issues.
In addition, the superior performance consists with our hypothesis that it is feasible and effective to recognize the facial expressions by a sequence of visual words from a global perspective.

\textbf{Results on CK+:} We also conduct cross-dataset evaluation to verify the superior generalization ability of our method.
Specifically, we first train the network individually on RAF-DB, FERPlus as well as AffectNet, and then evaluate the model directly on CK+.
Table \ref{ck_results} shows that our method achieves better performance than previous approaches. 
Note that gACNN and SPWFA-SE are trained for predicting seven basic emotions.
Although our model predicts one more expression (contempt), which generally lowers down the final results as shown in Table~\ref{aff_new}, the model trained on AffectNet gets an accuracy of 86.54\% on CK+.
Compared with the gACNN and the SPWFA-SE, our method has better performance and increases by 0.81\% and 0.16\%.
The model trained on AffectNet achieves higher accuracy than the ones trained on RAF-DB and FERPlus, because AffectNet is a relatively large scale dataset and contains more facial images.
Table \ref{ck_results} demonstrates that our method has better generalization capacity and achieves better performance without exceptions.

\subsection{Ablation Study}

As shown in Fig \ref{cvt}, our proposed method VTFF consists of LBP features with the attentional selective fusion (ASF) module and the multi-layer Transformer encoder (MTE).
To validate the effectiveness of these modules, we conduct comparative experiments on RAF, FERPlus and AffectNet by discarding some parts of our VTFF.
The detail settings of these experiments can are found in Table \ref{ablation_study}, where the setting (\textbf{a}) represents the baseline method.
Since the ASF is used to integrate the LBP features and the CNN features, it can not be retained without the LBP module.


\begin{table}[t]
  \caption{Ablation study w.r.t ASF and the concatenation \\fusion on RAF-DB, FERPlus, and AffectNet(mean $\pm$ standard deviation). $^{*}$ denotes the method using pretrained weights, otherwise the method is trained from scratch.}  \label{cat_asf}
  \centering
  \begin{tabular}{c|cccccc}
    \toprule
    Method  & RAF-DB & FERPlus & AffectNet\\ 
    \midrule
    Ours w/ Concat &  81.12 $\pm$ 0.25&  83.97 $\pm$ 0.37 &   58.07 $\pm$ 0.15\\
    Ours$^{*}$ w/ Concat &  87.35 $\pm$ 0.39  &  86.86 $\pm$ 0.16 &  59.85 $\pm$ 0.10\\
    \midrule
    Ours w/ ASF & 82.30 $\pm$ 0.31 &  84.71 $\pm$ 0.20 &  58.77 $\pm$ 0.09 \\
    Ours$^{*}$ w/ ASF & 87.83 $\pm$ 0.04 & 88.15 $\pm$ 0.23 & 60.90 $\pm$ 0.12 \\  
    \bottomrule
  \end{tabular}
\end{table}

\textbf{Effectiveness of the LBP features for FER.} Since our method begins with extracting the LBP features, we design the ablation study to investigate the impact of LBP features for FER.
Taking results on RAF-DB for example, the baseline without any modifications on RAF-DB achieves 86.37\% in terms of mean accuracy.
Adding the LBP features on RAF-DB improves the baseline by 0.16\%.
Settings (\textbf{a}, \textbf{b}) and settings (\textbf{d}, \textbf{e}) demonstrate that integrating the LBP features improves the baselines on FERPlus and AffectNet by 0.47\% and 0.20\%, which also suggests the LBP features are beneficial in improving expression recognition performance.
This can be explained by that the LBP features can extract texture information and reflect fine facial changes, which show the subtle differences of expressions.
Nevertheless, directly using additional the LBP features for FER is of limited use, because the simple addition fusion strategy is unsatisfactory for combining the LBP features and the CNN features.

\textbf{Evaluation of the attentional selective fusion (ASF).} 
To verify the effectiveness of the ASF for fusing LBP features and CNN features, we conduct experiments by replacing the ASF with the simple element-wise addition.
According to settings (\textbf{b}, \textbf{c}) and settings (\textbf{e}, \textbf{f}), the ASF leads to an increase in recognition accuracy when fusing the LBP features and the CNN features, showing the effectiveness of the proposed ASF.
Specifically, we can see from settings (\textbf{b}, \textbf{c}) that the designed ASF further improves the performance by 0.61\%, 0.38\% and 1.23\%.
According to the settings (\textbf{a}, \textbf{c}) the ASF improves the baseline setting (\textbf{a}) by 0.76\%, 0.85\% and 1.43\% with additional LBP features, respectively.
The ASF aggregates global and local contexts for fusing the LBP features and the CNN features, which further improves the recognition performance.

In addition, we compare our ASF with the concatenation operation.
Concatenating $L_{LBP}$ and $L_{RGB}$ is actually a data-level fusion method while our attentional selective fusion works at the feature-level.
We conduct corresponding experiments to verify whether the attentional selective fusion surpasses the concatenation operation.
To be specific, we add an extra layer to reduce the shape $H \times W \times 4$ into $H \times W \times 3$ to fit the input for the pretrained ResNet.
We fine-tune the whole model and also train the model from scratch.
From Table~\ref{cat_asf}, we can conclude that our method using the ASF significantly improves the performance on all datasets over that with the simple concatenation operation, which indicates that the feature-level fusion technique (i.e., ASF) surpasses the data-level fusion technique (i.e., concatenation) under this condition.

\begin{table}[t]
  \caption{Ablation study w.r.t. the number of heads, the number of layers, performed on RAF-DB, FERPlus and AffectNet.
  Bold values correspond to the best performance.}
  \centering
  \setlength{\tabcolsep}{1.2mm}{
\begin{tabular}{c|cccccc}
    \toprule
    Setting & $N_{l}$ & $N_{h}$ & Params(M) & RAF-DB & FERPlus & AffectNet\\
    \midrule
    \romannumeral1&4&4&51.8&87.45&88.46&61.08\\
    \romannumeral2 &4&8&51.8&\textbf{88.14}&88.69&61.55\\
    \romannumeral3 &4 &12 &51.8 &87.52&88.65&\textbf{61.85}\\
    \romannumeral4&8&4&80.1&87.22&88.14&60.82\\
    \romannumeral5 &8&8&80.1&87.61&\textbf{88.81}&61.23\\
    \romannumeral6 &8 &12 &80.1 &87.48&88.52 &61.30\\
    \romannumeral7&12&4&108.5&87.23&88.20&60.85\\
    \romannumeral8 &12&8&108.5&87.29&88.52&60.45\\
    \romannumeral9 &12&12&108.5&87.09&88.21&61.50\\
    \bottomrule
  \end{tabular}
  }
  \label{ablation_study2}
\end{table}

\begin{table}[t]
  \caption{Comparison between ViT and our VTFF. The p-values of these methods for RAF-DB, FERPlus and AffectNet are provided in the last two rows. The method marked with $^{*}$ denotes it inherits the pretrained weights, otherwise the method is trained from scratch.}
  \centering
  \setlength{\tabcolsep}{0.8mm}{
  \begin{tabular}{c|cccccc}
    \toprule
    Setting &$N_{l}$ & $N_{h}$ & Params(M)& RAF-DB & FERPlus & AffectNet\\
    \midrule
    ViT&12&12 &85.8 &47.55&47.72&27.87\\
    ViT$^{*}$&12&12 &85.8 &85.14&88.07&58.77\\
    \midrule
    Ours&4&8 &51.8&82.27&84.80&58.75\\
    Ours$^{*}$&4&8 &51.8&88.14&88.69&61.55\\
    \midrule
    p-values & - & - & - &$3.47 \times e^{-6}$&$7.30 \times e^{-6}$& $3.39 \times e^{-6}$\\
    p-values$^{*}$ & - & - & - & $6.68 \times e^{-5}$& $1.18\times e^{-4}$ & $2.57 \times e^{-5}$\\
    \bottomrule
  \end{tabular}
  }
  \label{ablation_study3}
\end{table}

\textbf{Effectiveness of the multi-layer Transformer encoder (MTE).} 
To explore the impact of the MTE, we evaluate the performance of the MTE in this part.
We also compare the effects of the MTE on RAF-DB, FERPlus and AffectNet.
It is worth to explain that we set the number of layers $N_{l} = 4$ and the number of heads $N_{h} = 8$ in Table \ref{ablation_study} to explore the effects of other modules and conduct a fair ablation study.
From settings (\textbf{a}, \textbf{d}), settings (\textbf{b}, \textbf{e}) and settings (\textbf{c}, \textbf{f}), we can clearly see that the MTE greatly improves the performance.
Specifically, compared with the baseline setting (\textbf{a}), the setting (\textbf{d}) shows that integrating the MTE outperforms the baseline by 1.23\%, 0.85\% and 2.05\% on RAF-DB, FERPlus and AffectNet.
Settings (a, b) and settings (b, c) demonstrate that integrating the LBP features and the ASF improves the baselines on RAF by 0.26\% and 0.61\% in terms of mean accuracy, respectively.
According to the settings (a, d), employing the MTE outperforms the baseline method ResNet18 by 1.23\%.
When keeping $N_{l}$ and $N_{h}$ constant in Table \ref{ablation_study}, we can conclude that the MTE contributes most to the accuracy improvements over the LBP features and the ASF.
In addition, we infer that employing the MTE increases the ability of learning discriminative features, outperforming corresponding baselines.

\textbf{Impact of the number of layers and heads of the MTE.} The multi-layer Transformer encoder consists of $N_{l}$ identical layers.
The multi-head self-attention in each layer enables the model decompose the information into $N_{h}$ representation subspaces and jointly capture discriminative information at different positions.
The effect of MTE module is dependent on its hyperparameters, i.e., $N_{l}$ and $N_{h}$.
Therefore, we investigate the effect of MTE with different hyperparameter setttings from 4 to 12 on RAF-DB, FERPlus and AffectNet in Table~\ref{ablation_study2}.
Table \ref{ablation_study2} compares the performance of our method in terms of accuracy/mean accuracy and parameters.
We observe that increasing the number of layers $N_{l}$ greatly burdens the parameters of the whole network.
Generally, smaller $N_{l}$ values tend to achieve better recognition performance, and larger $N_{l}$ values result in excessive parameters at the risk of overfitting.
Note that the smaller the value of $N_{h}$, the worse performance we may get, because there are not enough subspaces to learn latent representations.
Especially, it can be seen that the model produces the best performance on RAF dataset with $N_{l}=4$ and $N_{h}=8$.
 However, when $N_{l}=12$ and $N_{h}=12$, the MTE may even bring negative effect and the model produces worse recognition accuracy on RAF dataset than the model without MTE as shown in Table~\ref{ablation_study}. 
The reason is that the number of model parameters grows dramatically with the growth of  $N_{l}$ and $N_{h}$ and the 108.5M parameters of the model with $N_{l}=12$ and $N_{h}=12$ may be too many to be effectively optimized by learning from the small dataset like RAF. On the larger datasets like AffectNet, such negative effect of MTE is not observed.

\textbf{Impact of the pre-trained weights.} 
To find out the impact of pre-trained weights on our method, we train our model from scratch or fine-tune the model from pre-trained ResNet18 weights.
We also give a comparison of ViT and our VTFF in Table \ref{ablation_study3}.
We implement ViT-Base\cite{dosovitskiy2020image} on these three FER datasets and conduct experiments based on default settings as \cite{dosovitskiy2020image} described.
The ViT denotes that it is trained from scratch and ViT$^{*}$ represents we fine-tune it with weights pre-trained on ImageNet-21k and ImageNet.
From settings ViT$^{*}$ and Ours$^{*}$, we can infer that our method achieves better performance but with fewer parameters.
The performance of ViT greatly drops when trained from scratch instead of fine-tuning, because the feature extraction capacity of ViT is relatively limited without the guidance of large-scale datasets.
Table \ref{ablation_study3} shows that fine-tuning models from pre-trained weights usually results in better performance.
The training cost of our method for each dataset can be seen in Table \ref{train_cost}.
Theoretically, the training cost for the same method should be equal.
However, in practice, there exist slight differences between the experiments of the same method, because some other programs may allocate the GPU resource and the data I/O devices may work differently under various conditions.

\begin{table}[t]
  \caption{Training cost (GPU-days) of our \\method for RAF-DB, FERPlus and AffectNet.}\label{train_cost}
  \centering
  \setlength{\tabcolsep}{4mm}{
  \begin{tabular}{c|ccc}
    \toprule
    Method  & RAF-DB & FERPlus & AffectNet\\ 
    \midrule
    Ours & 0.06 & 0.07 & 0.36\\
    Ours$^{*}$ & 0.06 & 0.07 & 0.35\\    
    \bottomrule
  \end{tabular}
  }
\end{table}

\begin{table}[t]
  \caption{Impact of the ASF on ViT and our method. We run the experiment three times on each dataset with random seeds and compute the mean and standard deviation of the recognition accuracy.} \label{w_wo_asf}
  
  \centering
  \begin{tabular}{c|ccc}
    \toprule
     Method & RAF-DB & FERPlus &  AffectNet \\
    \midrule
     ViT$^{*}$ w/o ASF &  85.07 $\pm$ 0.30 &  87.93 $\pm$ 0.52 &  58.70 $\pm$ 0.09\\
     ViT$^{*}$ w/ ASF &  65.51 $\pm$ 0.39 & 69.03 $\pm$ 0.46 &  27.31 $\pm$ 0.18  \\
    \midrule
     Ours$^{*}$ w/o ASF & 87.83 $\pm$ 0.04 & 88.15 $\pm$ 0.23 & 60.90 $\pm$ 0.12 \\
     Ours$^{*}$ w/ ASF &  88.19 $\pm$ 0.21 &  88.70 $\pm$ 0.17 &  61.52 $\pm$ 0.07 \\
    \bottomrule
  \end{tabular}
\end{table}

We also provide a statistical evidence of the results to verify that the differences between them are statistically significant.
The McNemar's Test is used to compare the recognition accuracy of our proposed method and ViT.
We formulate the null hypothesis that the performance of the two models are equal. 
Thus, the alternative hypothesis is defined as: one of the two models performs better than the other.
For the assessment of the statistical significance, we set the significance threshold $\alpha=0.05$ to compute the p-values for all experiments.
The resulted p-values for all datasets are reported in Table \ref{ablation_study3}.
According to these results, we conclude that for all datasets the null hypothesis is rejected as the p-values were less than the significance level of 0.05, and thus, the differences between ViT and our method are statistically significant.
It also can be inferred from Table \ref{ablation_study3} that the improvements obtained by our model are not due to chance.

\textbf{Impact of the ASF on ViT and our VTFF.}
To investigate the impact of the ASF on ViT and our VTFF, we evaluate the methods with and without the ASF.
The experimental results are shown in Table~\ref{w_wo_asf}.
Under the same data settings, the vanilla ViT (ViT w/o ASF) does not work so well compared with our method (Ours w/ ASF) on all datasets.
We think the reason can be summarized as follows: a) Large numbers of parameters of ViT bring heavy burden on relatively small FER datasets when fine-tuing. To some extent, this is consistent with the ablation study on different layer values $N_{l}$ and the number of heads $N_{h}$. Larger $N_{l}$ and $N_{h}$ may result in worse performance.  b) The vanilla ViT is designed for FER-invariant tasks and takes nothing about FER into consideration, while our model integrates the ResNet pretrained on MS-Celeb-1M face recognition dataset and designs the ASF module for the feature fusion. Our method not only inherits the global representation capacity of Transformers, but also consists of task-specific modules for FER.

However, directly applying the ASF to fuse the inputs for ViT seems infeasible, because the ASF is a kind of feature-level fusion method.
To be specific, the ASF performs global and local feature fusion along the channel dimension.
The channel number of our method is 512, while the channel number of the inputs of ViT is 3.
The ASF may not be effective when fusing feature maps with such small channel number.
We train the ViT with the attentional selective fusion on the inputs to verify our hypothesis.
As we can see from the methods ViT w/o ASF and ViT w/ ASF in Table \ref{w_wo_asf}, the performances of ViT with the ASF drop dramatically compared with ViT without the ASF.
In addition, it can be concluded from the last two rows in Table \ref{w_wo_asf} that our model embedded with the ASF further improves the recognition performance, which demonstrates the effectiveness of the ASF for feature-level fusion.

\begin{figure}[t]
  \centering
    \centering
    \includegraphics[scale=0.58]{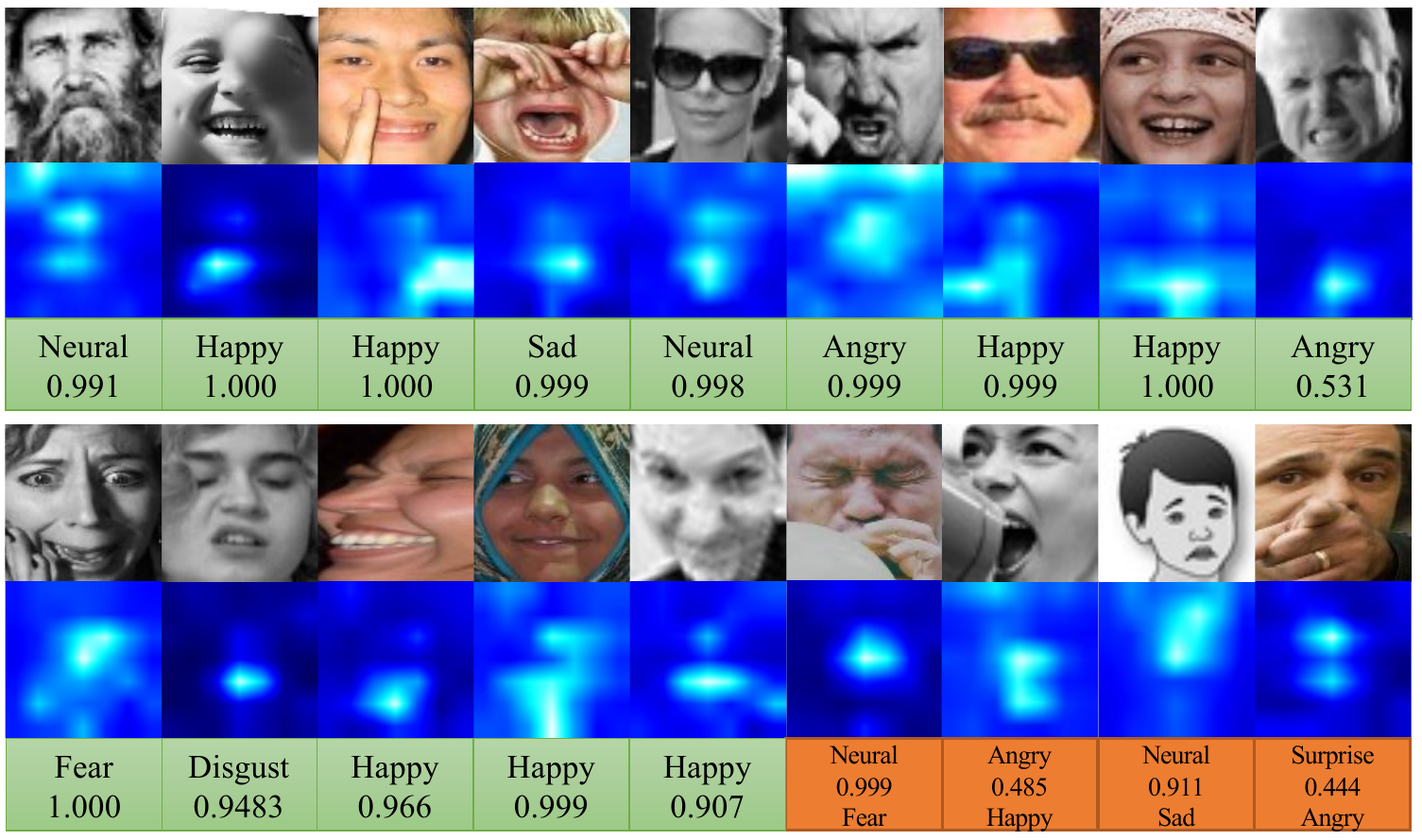}
\caption{An illustration of the learned attention maps. The first row is to show the raw images, the second row is the attention maps of the MTE, and the corresponding predictions or the labels are shown in the third row.
The annotations marked in green are the predicted expressions and their confidence scores, while the others in orange are false predictions with the ground-truth labels annotated.
}
\label{weightmap}
\end{figure}

\subsection{Visualization}
Our method computes relationships between visual words and captures discriminative features for expression recognition.
Table \ref{fer_results} provides an empirical evidence to support the effectiveness of our method.
To visually demonstrate the superior capacity of our method, the raw images and their corresponding attention weight maps are visualized in Fig. \ref{weightmap}.
We randomly select several images from RAF-DB, FERPlus and AffectNet, and compute the attention weight maps across all heads.
The weights of all layers are then multiplied recursively and projected into the input image space.

As shown in Fig. \ref{weightmap}, the weight maps shows that our method can identify occlusion regions and highlight discriminative features.
For example, the third raw image is occluded with a finger on the left part, while our method provides high attention weights on the right part.
The weight map of the eighth raw image contains high attention in the bottom, covering the mouth area.
We also provide some false predictions in Fig. \ref{weightmap}.
The first false-positive sample indicates that a man is blowing up a balloon about to burst and his eyes close out of fear.
It shows that the detection process is a double-edged sword, which provides the precious face locations but excludes the background knowledge.
The cartoon face is unfortunately predicted incorrectly because of limited cartoon faces in these datasets.
The other two false-positive samples also indicate the external knowledge like the social norms should be taken into consideration for the future research work. 
From Fig. \ref{weightmap}, We can conclude that our method can dynamically model the relationships among these visual words and highlight discriminative regions to boost recognition performance.

\section{Conclusion\label{conclusion}}
In this paper, we present the Visual Transformers with Feature Fusion (VTFF) for facial expression recognition in the wild.
We propose to tackle the expression recognition problem by translating facial images into sequences of visual words and performing recognition from a global perspective.
To achieve these goals, we design the attentional selective fusion to dynamically and adaptively combine LBP features and CNN features for improving recognition accuracy.
The visual words are generated by flattening and projecting the fused feature maps.
The multi-layer Transformer encoder utilizes the global self-attention mechanism to shift attention to discriminative visual words from a global perspective.
The experimental results demonstrate that the VTFF exceeds other state-of-the-art methods on three frequently used facial expression datasets, i.e., RAF-DB, FERPlus, AffectNet.
The cross-dataset evaluation on CK+ also demonstrates the promising generalization ability of our method.

\ifCLASSOPTIONcaptionsoff
  \newpage
\fi



%
\bibliographystyle{IEEEtran}
\bibliography{IEEEabrv,arxiv}

\end{document}